\documentclass[conference]{IEEEtran}
\usepackage{cite}
\usepackage{amsmath,amssymb,amsfonts}
\usepackage[ruled]{algorithm2e}
\SetKwInput{KwReturn}{Return}
\SetKwInput{KwDescription}{Description}

\usepackage{graphicx}
\usepackage{textcomp}
\def\BibTeX{{\rm B\kern-.05em{\sc i\kern-.025em b}\kern-.08em
    T\kern-.1667em\lower.7ex\hbox{E}\kern-.125emX}}

\usepackage{float} 
\usepackage[utf8, latin1]{inputenc} 
\usepackage{array}
\usepackage{fixltx2e}
\usepackage{stfloats}
\usepackage{url}
\usepackage{gensymb} 
\usepackage{multirow}

\usepackage{booktabs} 
\usepackage{gensymb}
\usepackage{xspace}
\usepackage{nicefrac}
\usepackage{xcolor, soul}
\usepackage[flushleft]{threeparttable}
\usepackage{caption}
\usepackage{subcaption}
\usepackage{tabularx}
\usepackage{tabularray}
\usepackage{cases}
\usepackage{siunitx}
\usepackage{color}

\DeclareMathAlphabet{\mathpzc}{OT1}{pzc}{m}{it}
    
\newcommand{\vect}[1]{\boldsymbol{\mathrm{#1}}}

\newcommand{\rv}[1]{\textcolor{black}{#1}}
\newcommand{\rvv}[1]{\textcolor{black}{#1}}

\makeatletter
\DeclareRobustCommand\onedot{\futurelet\@let@token\@onedot}
\def\@onedot{\ifx\@let@token.\else.\null\fi\xspace}

\def\eg{\emph{e.g}\onedot} 
\def\ie{\emph{i.e}\onedot} 

\def\etc{\emph{etc}\onedot} 

\makeatother

\IEEEoverridecommandlockouts
\usepackage{cite}
\usepackage{amsmath,amssymb,amsfonts}
\usepackage{algorithmic}
\usepackage{graphicx}
\usepackage{textcomp}
\usepackage{xcolor}
\def\BibTeX{{\rm B\kern-.05em{\sc i\kern-.025em b}\kern-.08em
    T\kern-.1667em\lower.7ex\hbox{E}\kern-.125emX}}

\begin{document}
\pagenumbering{arabic}

\title{Magazine supply optimization: a case-study
}


\author{Duong Nguyen, Ana Ulianovici, Sami Achour, Soline Aubry, and Nicolas Chesneau \\
\{firstname.lastname\}@ekimetrics.com
}

\maketitle
\thispagestyle{plain}
\pagestyle{plain}

\begin{abstract}

\rv{Supply optimization is a complex and challenging task in the magazine retail industry because of the fixed inventory assumption, irregular sales patterns, and varying product and point-of-sale characteristics. We introduce \textit{AthenIA}, an industrialized magazine supply optimization solution that plans the supply for over 20,000 points of sale in France. We modularize the supply planning process into a four-step pipeline: demand sensing, optimization, business rules, and operating. The core of the \rvv{solution} is a novel group conformalized quantile regression method that integrates domain expert insights, coupled with a supply optimization technique that balances the costs of out-of-stock against the costs of over-supply. \textit{AthenIA} has proven to be a valuable tool for magazine publishers, particularly in the context of evolving economic and ecological challenges.}

\end{abstract}

\section{Introduction}
\label{sec:introduction}

The print magazine has long been a staple in our culture, serving as a source of information, entertainment, and education. However, the rapid evolution of the Internet and digital media has profoundly transformed the media industry. In the last decades, there has been a marked shift in customer preferences toward digital platforms, which offer faster and more accessible content that appeals to the visual sensibilities of modern audiences. As newer generations increasingly favor instant online content over traditional print sources, print magazines are facing significant difficulties in maintaining both relevance and profitability \cite{kaiser_magazine_2012}. This ongoing shift has profound implications for print media companies, as they struggle to adapt to a landscape where digital media continues to gain domination. \rv{Moreover, the print magazine industry is confronting emerging sustainability challenges. Growing societal awareness of environmental impacts, shifting consumer preferences, rising production costs, and increasing sustainability-related taxes \cite{noauthor_definition_nodate, noauthor_plastics_nodate} have made sustainable practices within the industry essential.} 

In this context, the role of supply optimization becomes crucial, bearing significant impacts. On one hand, efficient supply helps increase profit by minimizing the losses due to out-of-stock (oos) and reducing the costs associated with excess supply. On the other hand, by curbing oversupply, the industry can mitigate its environmental footprint by reducing waste and the ecological ramifications of unsold print papers.

\rv{However, magazine supply optimization presents significant challenges. In magazine retail, the inventory is fixed. At the beginning of the selling period of an issue of a title, a predefined number of copies is sent to a specific list of Points Of Sale (POSes), referred to as the ``plate" in this paper. Replenishment is rare because the costs generally outweigh the benefits. At the end of the selling period, unsold copies are returned to the publisher, often to be destroyed. There are several layers of costs: production costs, registration costs (the costs of registering a title to the magazine retail network), distribution costs, and unsold costs (the costs related to the return and treatment-usually destruction-of unsold copies). The POSes also take commissions on the sold copies. To optimize the profit, the publisher must define which POSes to send magazine copies to (\ie, the plate of the issue) and estimate the optimal number of copies to send to each POS.
With the decline of the print media industry and the rise of environmental awareness, this optimization has become crucial, both economically and ecologically.
}

One of the key factors in supply optimization is the estimation of the demand, which encounters several other challenges: for each title, the content of one issue to another varies significantly, hence the demand. The demands are highly seasonal and greatly influenced by public holidays. Some POSes may be partially closed during the selling period. For new titles or when new POSes join the network, there are no historical data, leading to a cold-starting problem. Another challenge arises when issues are sold with ``extra-products" (a.k.a gift-on-purchase, they can be a toy model of a car, or a set of colored pencils, \etc.), which can significantly affect the demand. For example, an issue sold with a toy dinosaur typically sees a 10\% sales increase compared to normal.  In this paper, for simplicity, we use the term ``product" for the package sold to consumers, whether it is a copy of an issue alone or a copy of an issue with extra-product(s).

\begin{figure}
  \includegraphics[width=\linewidth]{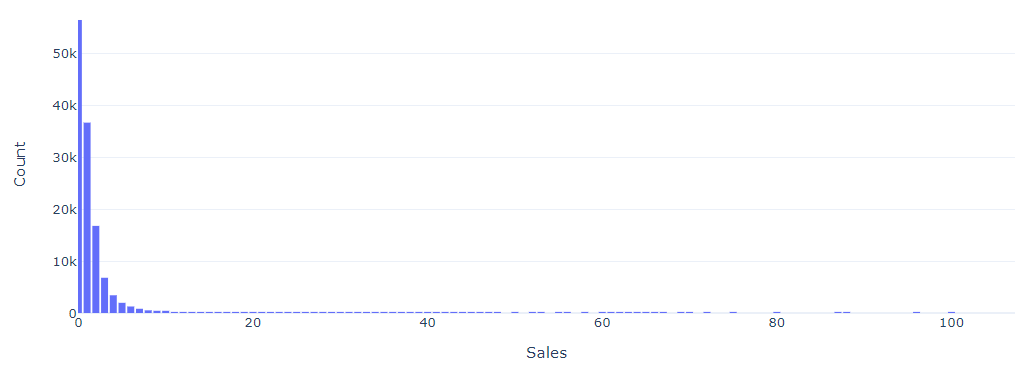}
  \centering
  \caption[]{Histogram of a random subset\footnotemark \textrm{ } of the sales. The sales follow a typical long-tail distribution.} 
  \label{fig:sales_histogram}
\end{figure}


\rvv{In magazine retail, magazines are sold through an extensive network of POSes (from a few thousand to tens of thousands). This vast network, combined with the high variability in sales, poses significant challenges. As shown in Figure \ref{fig:sales_histogram}, \footnotetext{We can not provide the histogram of the full data because of confidentiality reasons. The right hand side of the histogram was also cropped.}most POSes sell only 0, 1 or 2 copies of most issues, while a few POSes sell dozens to hundreds of copies of high-demand issues. Because of the large number of POSes, small errors in the estimation of the demand can accumulate across the plate, leading to substantial inefficiencies. Managing such a vast distribution network requires advanced methods capable of handling the inherent variability of magazine sales across numerous POSes.} 
\rv{Additionally, a supply optimization solution must adhere to various business constraints, such as maintaining a consistent presence of the title in the retail network. Furthermore, it must be flexible enough to accommodate different business strategies and potentially last-minute changes.}

\rv{To address these challenges, we developed \textit{AthenIA}---an industrialized data-centric magazine supply optimization solution. The solution was built for a magazine publisher in France. The publisher owns over 90 ``edutainment'' (education and entertainment) magazine titles, which vary in periodicity, including weekly, monthly, bimonthly, quarterly, and special editions. The magazines are sold in a highly regulated retail network comprising more than 20,000 POSes. 
The key points of the solution are as follows:}
\begin{itemize}
    \item \rvv{Data-centric pipeline for supply optimization: we introduce a data-centric pipeline that modularizes the supply optimization process into four stages. This approach facilitates the design, evaluation, method selection, and maintenance of each stage. The pipeline is both methodological and agile, enabling it to adapt to various scenarios and supply strategies. Its design allows for generalization to other sectors beyond magazine publishing.}
    \item \rv{Human-in-the-loop group conformalized quantile demand sensing: we introduce a group conformalized quantile demand sensing approach that enhances the integration of domain expert knowledge to model different nuances in the characteristics of the products.}
    \item  \rvv{Novel optimization technique: we present a novel optimization technique that optimizes the profit across a retail network with an extensive number of POSes, under various constraints, given the estimated demand and the nonlinear and complex cost structure of the business.}
\end{itemize}

\rv{The paper is organized as follows: We formulate the problem in Section \ref{sec:problem_formulation}. Section \ref{sec:solution} provides a detailed description of the solution, followed by an analysis of the results in Section \ref{sec:evaluation}. We outline the deployment details in Section \ref{sec:deployment}, and compare the solution with related work in Section \ref{sec:related_work}. Finally, Section \ref{sec:conclusion} concludes with a discussion regarding the impact of the solution and practical considerations.}

\section{Problem formulation}
\label{sec:problem_formulation}

Let us denote $<m,i>$ issue $i$ of title $m$. Mathematically, the objective of the solution is to optimize the profit under constraints, as follows:
\begin{equation}
\label{eq:obj_global}
\begin{aligned}
\max_{s^{j}_{m,i}} \quad & \sum_{j \in \mathbb{P}^{}_{m, i}} \left( z^{j}_{m,i}*{price}_{m, i} - {cost}^{j}_{m, i}(s^{j}_{m,i}, s_{m,i}, z^{j}_{m,i}) \right)\\
\textrm{s.t.} \quad & s^{j}_{m,i}, z^{j}_{m,i} \in \mathbb{Z}^{0+},\\
  & s_{m,i} \in \left[N_{total}-\Delta, N_{total}+\Delta \right].
\end{aligned}
\end{equation}
with 
\begin{itemize}
    \item $\mathbb{P}^{}_{m, i}$ denotes the plate of $<m,i>$,
    \item $s^{j}_{m,i}$ denotes the supply of $<m,i>$ for POS $j$ (\ie how many copies of $<m,i>$ sent to POS $j$),
    \item $\mathbb{Z}^{0+}$ denotes the set of non-negative integers, 
    \item $s_{m,i} \triangleq \sum_{j \in \mathbb{P}^{}_{m, i}} s^{j}_{m,i} $ denotes the total supply of $< m,i >$,
    \item $N_{total}$ denotes the total supply constraint of $<m,i>$, imposed by the publisher,
    \item $\Delta$ denotes the tolerance of the total supply constraint,
    \item $z^{j}_{m,i}$ denotes the sales of $<m,i>$ at POS $j$,
    \item ${price}_{m, i}$ denotes the price of $<m,i>$,
    \item ${cost}^{j}_{m, i}(s^{j}_{m,i}, s_{m,i}, z^{j}_{m,i})$ denotes the costs and commission taken by POS $j$ when it is supplied $s^{j}_{m,i}$ of $s_{m,i}$ total copies, and it sells $z^{j}_{m,i}$ copies.
\end{itemize}
\rv{In layman's terms, Eq. \eqref{eq:obj_global} means to \textit{``find the number of copies sent to each POS in plate $\mathbb{P}^{}_{m, i}$ to maximize the profit of issue $i$ of title $m$, given the cost structure ${cost}^{j}_{m, i}$ and the total supply constraint $N \pm \Delta$''}}. 

In Eq. \eqref{eq:obj_global}, ${price}_{m, i}, N_{total},$ and $\Delta$ are known. $\mathbb{P}^{}_{m, i}$ is provided by a set of business rules to maintain the presence of the title and to conform to the regulations of the retail network.  The formula of ${cost}^{j}_{m, i}$ is also known. However, the value of ${cost}^{j}_{m, i}(s^{j}_{m,i}, s_{m,i}, z^{j}_{m,i})$ depends not only on $s^{j}_{m,i}$ and $z^{j}_{m,i}$ but also on the total supply $s_{m,i}$. Another difficulty of Eq. \eqref{eq:obj_global} is the sales $z^{j}_{m,i}$ depends on the supply  $s^{j}_{m,i}$. Note that $z^{j}_{m,i}$ denotes the sales, not the demand. Let us denote $d^{j}_{m,i} \in \mathbb{R}^{0+}$ ($\mathbb{R}^{0+}$ denotes the set of non-negative real numbers) the demand of $<m,i>$ at POS $j$. We have: 
\begin{equation}
\label{eq:sales_demand}
    z^{j}_{m,i} = \min(d^{j}_{m,i}, s^{j}_{m,i}).
\end{equation}
When $s^{j}_{m,i} > z^{j}_{m,i}$, the sales are equal to the demand. However, when there is an out-of-stock (\ie $s^{j}_{m,i} = z^{j}_{m,i}$), there is potentially an unobserved demand that was missing. We can measure only the sales, not the actual demand. Moreover, the demand also depends on the supply. For instance, suppose that we send 4 copies to a POS and it sells 3. This does not imply that if we sent 10 copies to this POS, it would sell the same 3. This ``visibility effect" is common in retail \cite{gul_retail_2021, lu_developing_2015}. 

We detail in the next section how \textit{AthenIA} addresses these challenges. 
\section{AthenIA supply optimization solution}
\label{sec:solution}
The variability in the products' characteristics and in the demand poses significant challenges for building a fully automatic supply optimization solution. A purely data-driven approach falls short of capturing all the nuances of the products.  
To address this, we built a data-centric solution that not only leverages patterns in the data but also integrates insights of Subjet-Matter-Experts (SMEs) and accommodates various business constraints and publisher's operational strategies. This approach ensures that the solution is not only theoretically optimal but also practically useful and usable. The details will be presented in the subsequent sections. We limit the scope of this paper to products that have historical data\footnote{When the publisher publishes a new title, a different supply strategy is used to promote the visibility of the product. This is out of the scope of this paper.}. 

\subsection{Pipeline}
\label{subsec:pipeline}
\begin{figure}
  \centering
  \includegraphics[width=\linewidth]{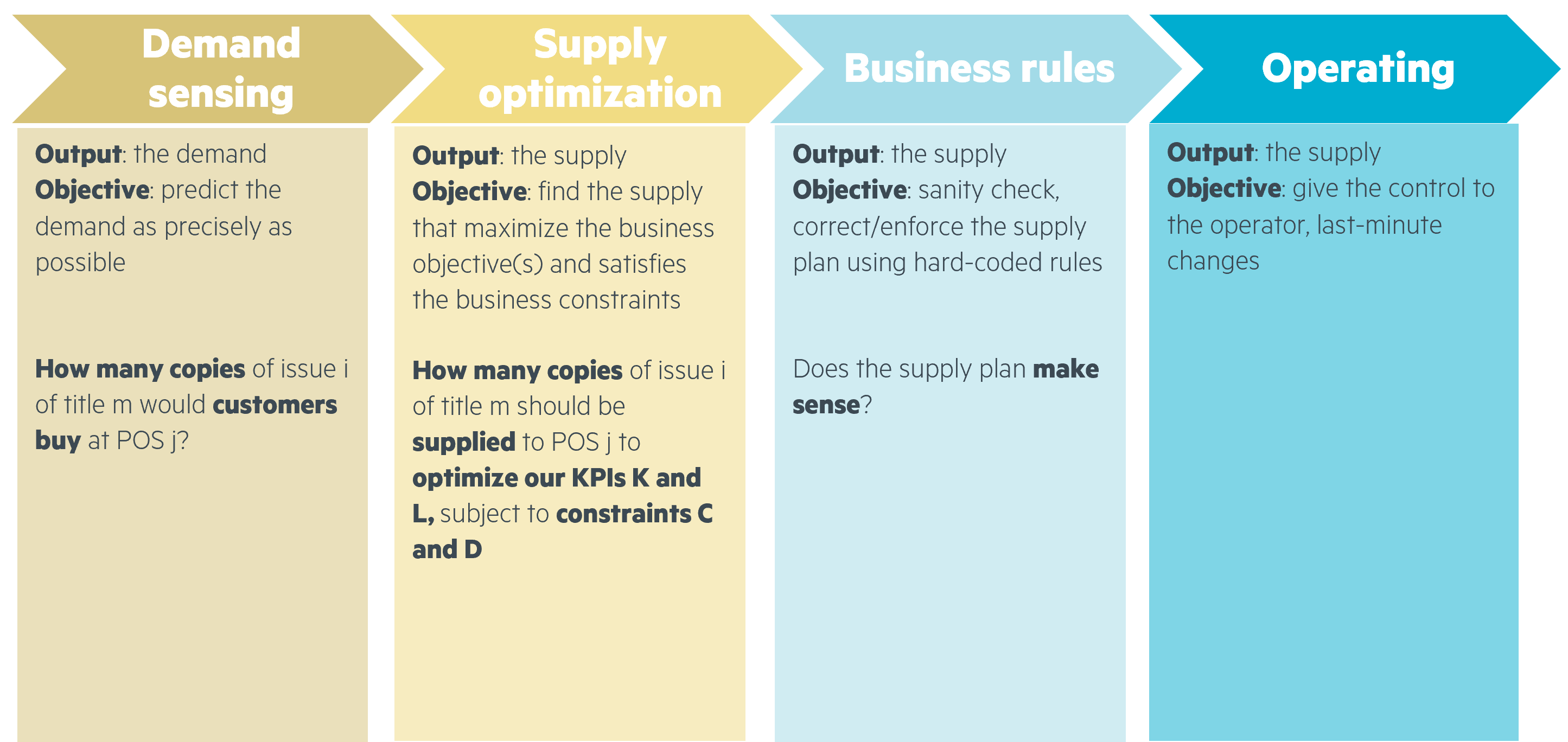}
  \centering
  \caption{The supply optimization process is decomposed into four stages.} \label{fig:pipeline}
\end{figure}

We modularize the supply optimization process into four stages, as shown in Figure  \ref{fig:pipeline}.
\begin{itemize}
  \item Stage 1 (Demand sensing): at this initial stage, we estimate the demand for each issue at each POS. It is important to note that we are predicting the demand, not the sales. This estimation is at the heart of the whole process.
  \item Stage 2 (Optimization): based on the demand estimated from Stage 1, we calculate the optimal supply for various scenarios, taking into account different costs, constraints, and strategies.
  \item Stage 3 (Business rules): at this stage, we apply some predefined rules to sanity-check the proposed supply plans and make ad-hoc adjustments to adapt to unplanned changes.
  \item Stage 4 (Operating): the solution generates several optimal supply plans, each corresponding to a specific strategy. The supply planner then selects the most appropriate plan for each issue and makes any necessary manual adjustments.   
\end{itemize}

Modularizing the supply planning process into several sub-stages provides several benefits:
\begin{itemize}
    \item Impact measurement and decision making: by breaking the process into smaller stages, we can measure the impact of each step individually. This allows us to identify which stages add value and which need improvement.
    \item Flexibility: in our pipeline, from left to right the process changes from highly automatic, data-influenced to manual, domain-expertise-influenced. This design allows users to bring in the more up-to-date knowledge of the products and the nuances that can not be fully captured by an automatic model. This flexibility is advantageous for operational utility. When market conditions, regulations, or cost structures change, we can easily adapt specific stages of the pipeline, especially the later stages (operating, business rules), without overhauling the entire system, thereby reducing update and maintenance costs.   
\end{itemize}

The following sections will provide detailed descriptions of each stage.

\subsection{Human-in-the-loop demand sensing using group conformalized quantile regression}
\label{subsec:demand_sensing}

The demand of issue $i$ of title $m$ at POS $j$ depends on several factors: the quality, the popularity, and the trends of the title; the quality of this particular issue; the extra-product (if any); the market trends; the selling period; the characteristics of the POS; the visibility of the product on the shelf; \rvv{the sales of concurrent products}; among others. Although $d^{j}_{m,i}$ is also influenced by the supply $s^{j}_{m,i}$, incorporating this dependence into the modeling would make the problem intractable. Moreover, we can not observe the demand directly, we can only observe the sales. When an out-of-stock situation occurs, the actual demand may exceed the observed sales. To work around these complexities, we make the following simplifications:

\begin{itemize}
    \item \textit{Simplification 1}: if the supply exceeds the sales, the demand is considered equal to the sales. Conversely, in cases of out-of-stock, we assume that the POS could have sold an additional $r$\%:
    \begin{equation}
    d^{j}_{m,i}=\begin{cases}
      z^{j}_{m,i} & \textrm{, if } z^{j}_{m,i} < s^{j}_{m,i}\\
      (100+r)\%*z^{j}_{m,i} & \textrm{, if } z^{j}_{m,i} = s^{j}_{m,i}
    \end{cases}
    \end{equation}
    \item \textit{Simplification 2}: $d^{j}_{m,i}$ does not depend on $s^{j}_{m,i}$
\end{itemize} 

\textit{Simplification 1} allows us to approximate the demands of historical issues using observed sales data, while \textit{Simplification 2} makes the supply optimization problem tractable.

\begin{figure}
  \includegraphics[width=\linewidth]{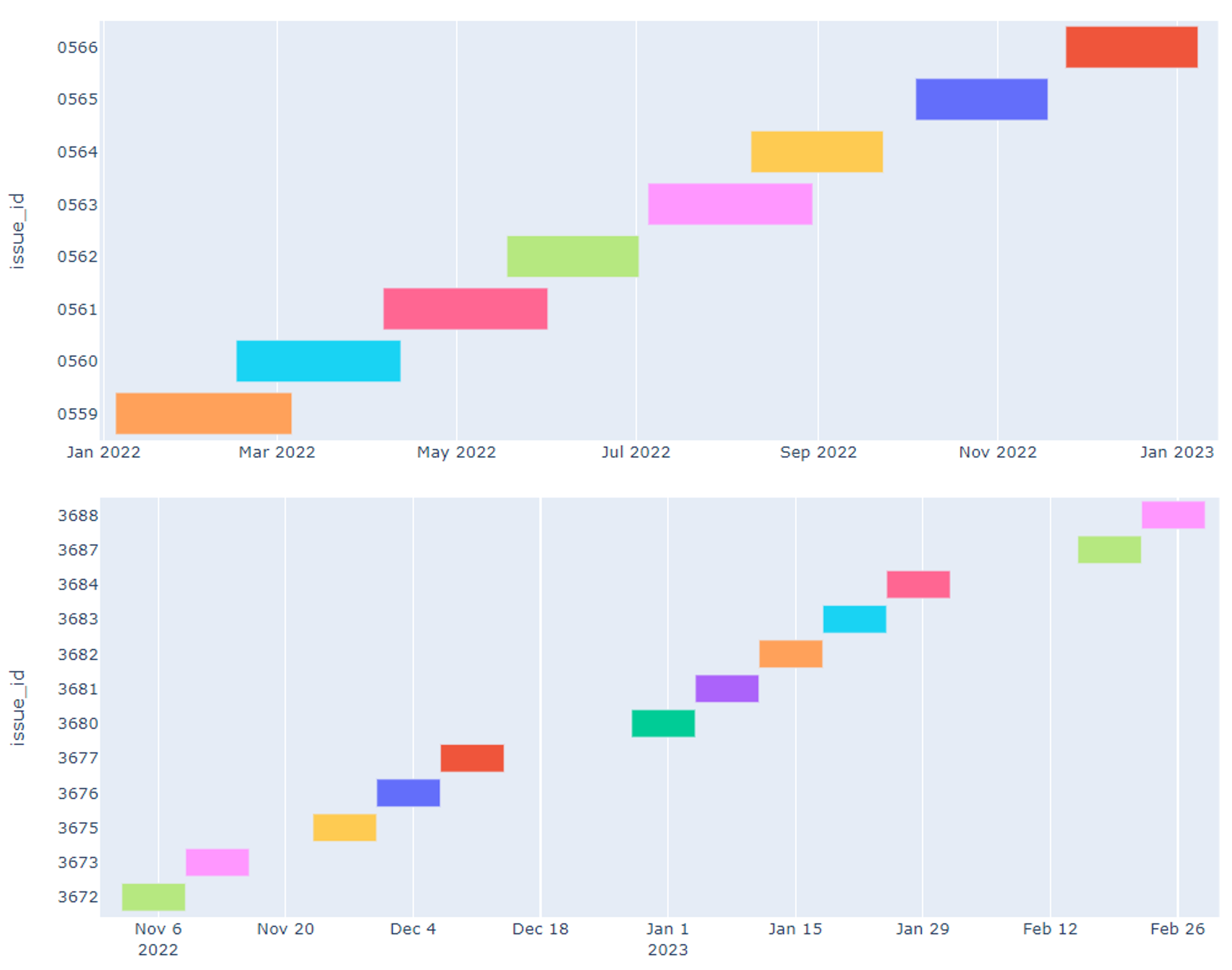}
  \centering
  \caption{The irregularity in the sales of magazines.} 
  \label{fig:sale_period}
\end{figure}

We build a demand sensing model to estimate $d^{j}_{m,i}$ using historical demands, product characteristics, sales characteristics, and contextual information. The irregularity in the sales of the magazines limits the possible choices of models for this task. Figure \ref{fig:sale_period} illustrates the selling periods of two titles in the dataset. The selling start dates of the issues are irregular, and the duration of each issue's selling period varies (top). Additionally, the publisher may skip one or several issues (bottom).  Common time series models requiring regularly sampled data \cite{salinas_deepar_2019, oreshkin_n-beats_2020, lim_temporal_2020} are not applicable.  Therefore, we convert the data into a tabular format and use LightGBM \cite{ke_lightgbm_2017} for our demand sensing model. This approach allows us to handle irregular selling periods and varying issue durations effectively. Similar architectures, such as CatBoost \cite{prokhorenkova_catboost_2018}, are also an option and give similar results (not reported in this paper). 

\textbf{Human-in-the-loop feature engineering}: common feature engineering to extract the characteristics of the historical demands includes using the trends and the lag values to capture the trend and the seasonality, respectively. We use the three most recent observed issues of the same title for the trend. Given that the seasonality in magazine retail is yearly, we use the 1-year and 2-year lags of the same title. Additionally, we calculate the average sales over 12 months and 6 months, lagged 3 months. 

\begin{figure}
  \includegraphics[width=\linewidth]{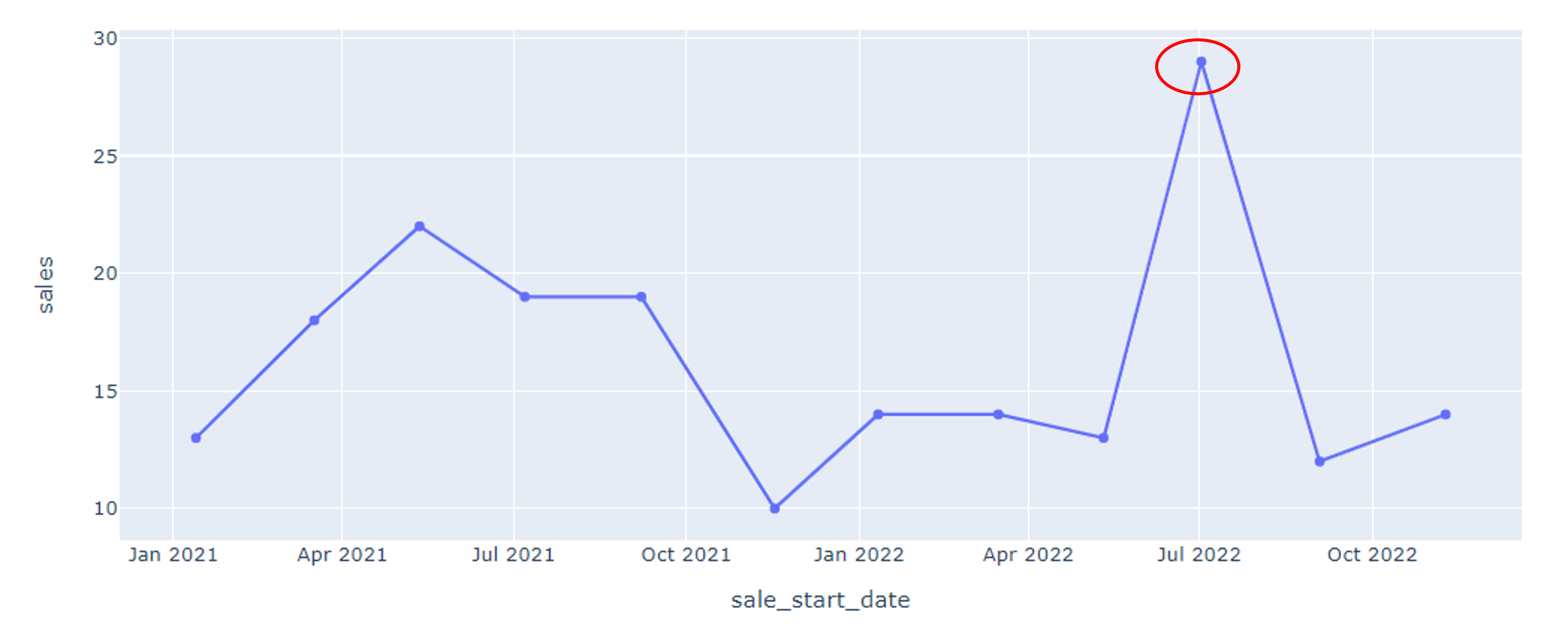}
  \centering
  \caption{Atypical sales. The circled peak was because that issue was sold with a particularly successful extra-product.} 
  \label{fig:abnormal_sales}
\end{figure}

The demand is significantly affected by the variations between issues, such as differences in appearance, packaging, and content, as well as the inclusion of extra-products. For example, the atypical peak in Figure  \ref{fig:abnormal_sales} was due to a particularly successful extra-product, which disrupted the calculation of trend and seasonality patterns. To mitigate this, we allow SMEs to alter the trend and seasonality issues by excluding atypical issues in the calculation. Furthermore, for each issue, the SMEs assign two ``references", which are two similar issues in the historical data (not necessarily of the same title).  This human-in-the-loop approach helps the solution capture the subtle nuances in the characteristics of the products. However, there is a potential risk of human bias. We accept this trade-off for the valuable insights provided by SMEs. The impact of the references will be analyzed in the next section.

Given that there are more than three hundred different extra-products, using one-hot encoding for each would be inefficient. Instead, we categorize these extra-products into eight groups based on their financial performance, which is defined as the average per-unit profit (before tax) of the issues utilizing that extra-product in the historical data. We then apply ordinal encoding to these eight categories.


To address the high disparity in the sales (and the demand), as shown in Figure \ref{fig:sales_histogram}, during the training phase, we convert the target variable (the demand) to a log scale \cite{laptev_time-series_2017, kunz_deep_2023}. The impact of this scale transformation will be analyzed later in this paper. 

Table \ref{tab:feature_engineering} in the Appendix describes all the features used in the demand sensing model.

\textbf{Group Conformalized Quantile Regression (GCQR)}: In supply optimization, the costs associated with oversupply and undersupply are different. Quantile regression \cite{koenker_regression_1978} is a commonly used approach in these cases. Recall that the $\alpha$th conditional quantile function is defined as:
\begin{equation}
    q_{\alpha}(\vect{x}^j_{m,i}) \triangleq \inf \{ d:Pr[D^j_{m,j} \leq d \textrm{ }| \textrm{ } \vect{x}^j_{m,i}] \geq \alpha \}
    \label{eq:quantile}
\end{equation}
with $D^j_{m,j}$ denotes the random variable of the demand and $\vect{x}^j_{m,i}$ denotes the input features. Conditional quantile regression estimates $q_{\alpha}(\vect{x}^j_{m,i})$ using a regression model with the following pinball loss function:
\begin{equation}
    loss_{\alpha}(d, \hat{d})=\begin{cases}
      {\alpha}*(d - \hat{d}) & \textrm{, if } \hat{d} <= d\\
      (1-{\alpha})*(\hat{d} - d) & \textrm{, if } \hat{d} > d
    \end{cases}
    \label{eq:pinball}
\end{equation}
with $d$ and $\hat{d}$ denote the true target value and the predicted value of the regression, respectively. However, quantile regression often gives prediction intervals that are too wide \cite{romano_conformalized_2019}. \rv{As a result, using standard quantile regression for demand sensing in supply optimization often leads to oversupply. To address this problem, Conformalized Quantile Prediction (CQP) \cite{shafer_tutorial_2008, romano_conformalized_2019, angelopoulos_gentle_2022} uses a calibration set to calibrate the outputs of quantile regression, resulting in more accurate and narrower prediction intervals. Based on this idea, we designed a new group conformalized quantile regression, tailored to our problem, as follows:}
\begin{itemize}
    \item Step 1 (Split): for each title $m$, we split the historical data into two sets: a calibration set $\mathit{Cal}_{m}$ containing the observations of the two latest observed issues if the title's periodicity is more frequent than quarterly (\eg weekly, monthly, bimonthly), otherwise the lastest observed issue; and a training set $\mathit{Tr}_{m}$ containing the rest of the historical data (which means there are not only the observations of title $m$ but also those of other titles in $\mathit{Tr}_{m}$). 
    \item Step 2 (Training): \rv{we train a quantile LightGBM regression model $\mathcal{M}_{m,\alpha}$ using the data in the training set $\mathit{Tr}_{m}$.}
    \item Step 3 (Calibration): we apply $\mathcal{M}_{m,\alpha}$ on $\mathit{Cal}_{m}$. Let us denote $e^j_{m,i,\alpha}|_{\mathit{Cal}_{m}}$ the prediction error of $\mathcal{M}_{m,\alpha}$ at POS $j$ for issue $i$ of title $m$ in the calibration set $\mathit{Cal}_{m}$. 
    \item Step 4 (Grouping): \rv{we split the observations in all the calibration sets into $G$ groups $\{G_1, G_2, ..., G_G \}$ based on the average sales over the last 12 months lagged 3 months (\texttt{mean\_sales\_12m}).}
    \item Step 5 (Correction): for each group $G_g$, we calculate the $\alpha$th quantile $q^{g}_{\alpha}$ of the set of the calibration prediction errors $\{ e^j_{m,i,\alpha}|_{\mathit{Cal}_{m}} \}$ of the observations in this group. 
    \item Step 6 (Prediction): \rv{for each observation in the test set, its group is determined using the same rule as in Step 4. The final $\alpha$th quantile prediction of this observation is $\hat{d}^j_{m,i} \triangleq \mathcal{M}_{m,\alpha}(\vect{x}^j_{m,i}) + q^{g}_{\alpha}$}.
\end{itemize}

The splitting in Step 4 is to reduce the disparity in the scale of the observations. For instance, the correction term $q^{g}_{\alpha}$ for a group that has \texttt{mean\_sales\_12m} between 0 and 1 should differ from the correction term for the group that has \texttt{mean\_sales\_12m} between 50 and 100 (see Figure \ref{fig:correction_terms} for an illustration of how these correction terms vary across different groups). 
\subsection{Supply optimization}
\label{subsec:supply_optimization}

\begingroup
\begin{table}[htbp]
    \centering
    \begin{tblr}{
      colspec={|X|X|X|X|X|}, hlines, 
    }
      Combination & Group $G_1$ & Group $G_2$ & ... & Group $G_G$ \\
      Comb 1 & 0.65 & 0.65 & ... & 0.70 \\
      Comb 2 & 0.65 & 0.65 & ... & 0.75 \\
      ... & ... & ... & ... & ... \\
      Comb C & 0.90 & 0.95 & ... & 0.99 \\
    \end{tblr}
    \caption{\raggedright Scenarios table. Each row corresponds to a combination to test, each column corresponds to a group. The value of each cell is $\alpha$.}
    \label{tab:scenarios}
\end{table}
\endgroup

The demand sensing model in Stage 1 gives us the prediction $\hat{d}^j_{m,i,\alpha}$ of the $\alpha$th quantile of the demand. If we use $\hat{d}^j_{m,i,\alpha}$ as the supply, the out-of-stock rate is expected to be $1-\alpha$. However, there are several problems. Firstly, $\hat{d}^j_{m,i,\alpha}$ is a real number, whereas the supply $s^j_{m,i}$ must be a non-negative integer. Given that the majority of the sales are 0, 1, or 2 copies, the rounding residuals will negate the precision of the demand sensing model. For example, rounding up $\hat{d}^j_{m,i,\alpha} = 0.8$ to 1 introduces a rounding residual of 0.2, which is 25\% of  $\hat{d}^j_{m,i,\alpha}$. Secondly, using a fixed rule to map $\hat{d}^j_{m,i,\alpha}$ to $s^j_{m,i}$ does not guarantee that the sum of $s^j_{m,i}$ across all the POSes in the plate $\sum_{j \in \mathbb{P}^{}_{m, i}}s^j_{m,i}$ will fall within the supply constraint interval $N_{total} \pm \Delta$. One may increase or decrease proportionally $s^j_{m,i}$ to meet the supply constraint. However, that means we undermine the efforts done in the demand sensing step. Another possible solution is to sweep $\alpha$ in the $\alpha$th quantile regression over different values until the sum of the supply meets the constraint. However, this approach does not guarantee that the final profit will be optimized, since the costs are not taken into account. 

\begin{figure}
  \centering
  \includegraphics[width=\linewidth]{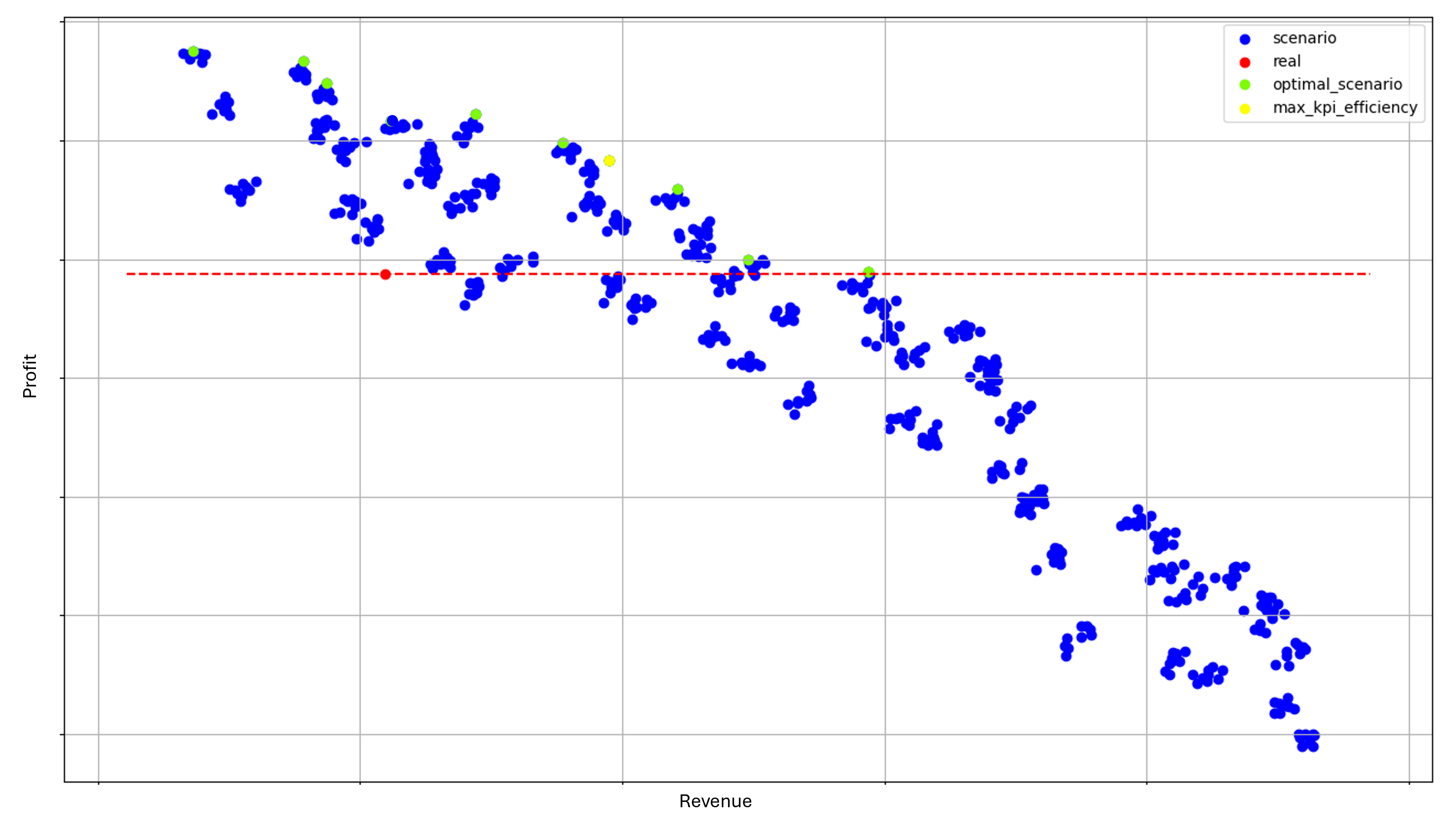}
  \centering
  \caption{Example of the revenue and profit of different scenarios. Each dot corresponds to a scenario. See text for details.}.  \label{fig:optim_result}
\end{figure}

To address these challenges, we design a new technique to optimize the supply for each $<m, i>$ as follows:
\begin{itemize}
    \item Step 1: instead of estimating one quantile $\hat{d}^j_{m,i,\alpha}$, we estimate several quantiles $\hat{d}^j_{m,i,\alpha}$ corresponding to different values of $\alpha \in \{0.65, 0.70, 0.75, 0.80, 0.85, 0.90, 0.95, 0.99\}$. 
    \item Step 2: we construct several ``scenarios", each scenario is represented by a vector $\left[\alpha_1, \alpha_2,... ,\alpha_G \right]^T$ where $\alpha_g$ corresponds to the value of $\alpha$ for group $G_g$. An example is shown in Table. \ref{tab:scenarios}. The number of scenarios to test is a trade-off between the exhaustiveness and the calculation costs.
    \item Step 3: for each scenario, we obtain the $\alpha$th quantile predictions $\hat{d}^j_{m,i,\alpha}$ of the observations in the calibration set $\mathit{Cal}_m$, and define the scenario supply $\hat{s}^j_{m,i}$ as the ceiling of $\hat{d}^j_{m,i,\alpha}$. The corresponding estimated sales are defined as the minimum between the real observed sales and the scenario supply: $\hat{z}^j_{m,i} \triangleq \min(z^j_{m,i}, \hat{s}^j_{m,i})$. 
    \item Step 4: \rvv{given the scenario supply, the corresponding sales, and the cost structure, we calculate business KPIs (Key Performance Indicators): total supply, revenue, costs, and profit for each scenario.}
    \item Step 5: We compare the KPIs of the scenarios with the real KPIs and select a set of ``optimal" scenarios (the green dots in Figure \ref{fig:optim_result}). The criteria for selecting the optimal scenarios are predefined by the business stakeholders. These criteria are mainly based on the total supply, the revenue, and the profit. The best optimal scenario is referred to as \textit{max\_kpi\_efficiency}. 
    \item Step 6: During the inference phase, for each issue $<m, i>$, we calculate the supply corresponding to each scenario in the set of optimal scenarios obtained in Step 5. \textit{AthenIA} outputs three supply plans. The first plan, dubbed as the \textit{optimal supply plan} corresponds to the \textit{max\_kpi\_efficiency} scenario. The second plan, dubbed as the \textit{optimal distribution plan} is the supply plan corresponding to the optimal scenario that gives the total supply closest to the constraint $N_{total}$. The third supply plan, not reported in this paper, uses a similar supply optimization strategy as the \textit{optimal supply plan's} but on a larger plate than $\mathbb{P}_{m,i}$. This plan is used when the publisher wants to widen the presence of the magazine. 
\end{itemize}

This method enables \textit{AthenIA} to accommodate complex cost structures and business strategies. The \textit{optimal supply plan} gives us the supply plan that would maximize the profit, while the \textit{optimal distribution plan} gives the plan that would optimize the distribution of the predefined total supply among the POSes in the plate. Occasionally, the \textit{optimal supply plan} and the \textit{optimal distribution plan} can be identical. 

\subsection{Business rules and Operating}
\label{subsec:business_rules}

We apply a set of rule-based checks and adjustments to control the outputs of Stage 2. These rules serve two purposes:
\begin{itemize}
    \item Sanity check: since Stages 1 and 2 are run automatically using machine learning models, it is essential to apply sanity checks to verify whether the proposed supply of each POS is too high or too low compared to historical trends and yearly lags.
    \item Fast adjustment: \rvv{When the supply planners acquire new market insights that need to be incorporated into the supply plans, or when they want to apply some new supply strategies for atypical products}, rule-based adjustments offer a quick and efficient way to implement the planners' intentions. Modifying the demand sensing model or the optimization strategies would be more time-consuming and require rigorous testing.   
    \item Constraint adjustment: Occasionally, none of the plans provided by \textit{AthenIA} meets the total supply constraint $N_{total} \pm \Delta$. In these rare instances, we review with the supply planner to decide whether to adjust the supply constraint or to scale the supply proportionally to meet the constraint. This situation is infrequent because the range of the total supplies from different optimal scenarios is extensive, and the supply planners typically have sufficient prior knowledge of their products' sales to set reasonable total supply constraints.
\end{itemize}
These rule-based checks and adjustments ensure that the final supply plans are not only optimized but also practical and aligned with current market conditions and constraints.

The last stage in the process is called the Operating stage.  At this stage, the supply planners review the three supply plans generated by \textit{AthenIA} and select the most suitable one. If necessary, the planners make ad-hoc adjustments to better align with current conditions or unforeseen changes. Once the adjustments are made, the finalized supply plan is sent to the production and shipping departments for execution. \textit{AthenIA} logs the planners' choice and any modifications made for future analysis.

\section{Evaluation}
\label{sec:evaluation}

The ultimate goal of the solution is to optimize the profit within given constraints. However, due to the multi-stage nature of the process, it is essential to evaluate the relevance and the performance of each stage individually. Without this evaluation, the number of potential combinations of options for each stage becomes unmanageably high. By assessing each stage separately, we can ensure that each component contributes effectively to the overall optimization, thereby streamlining the process and enhancing the solution's efficiency. This approach allows us to identify and fine-tune the critical elements that drive the most significant impact on profit optimization, ensuring that our methodology remains both robust and practical.

\subsection{Evaluation of the demand sensing model}
\label{subsec:eval_demand_sensing}

To evaluate the performance of the demand sensing model, we tested it on a dataset containing sales records of issues put on sale from 01/01/2021 to 31/05/2023. We simulated the situation as of 01/04/2023: we used all the sales records available on that date to train the model, then predicted the demands of issues put on sale from 01/04/2023 to 31/05/2023. 
This setup is intended solely to facilitate the reproducibility of the results presented in this paper. In operation, we retrain the model on a weekly basis with the latest data.

We compared the performance of the \textit{AthenIA} demand sensing model with several baselines and alternative options:
\begin{itemize}
    \item Naive: The latest observation is used as the prediction. 
    \item S-Naive: The 1-year lag is used as the prediction.
    \item LightGBM\_standard: A LightGBM model that uses the features in Table \ref{tab:feature_engineering}, excluding  \texttt{mean\_ref} and  \texttt{max\_ref}, as inputs. The target variable is transformed onto a log scale during the training phase.
    \item LightGBM\_linear: A LightGBM that uses all the features in Table \ref{tab:feature_engineering} as inputs. However, the target variable is NOT transformed onto a log scale during the training phase. 
    \item LightGBM\_log:  A LightGBM that uses all the features in Table \ref{tab:feature_engineering} as the inputs. The target variable is transformed onto a log scale during the training phase.
    \item LightGBM\_log\_GCQR (\textit{AthenIA}): the LightGBM\_log model with GCQR correction. This is the model used in production. 
\end{itemize}

We set $r\%$ = 30\% and compared the performance of different models using the mean pinball loss (Eq. \eqref{eq:pinball}) on the test set with various values of $\alpha$. Although we cannot guarantee that the model with the lowest pinball loss will always be the best model for the downstream supply optimization task, the pinball loss remains the most suitable evaluation metric for our purposes. 

\begin{figure}
  \centering
  \includegraphics[width=\linewidth]{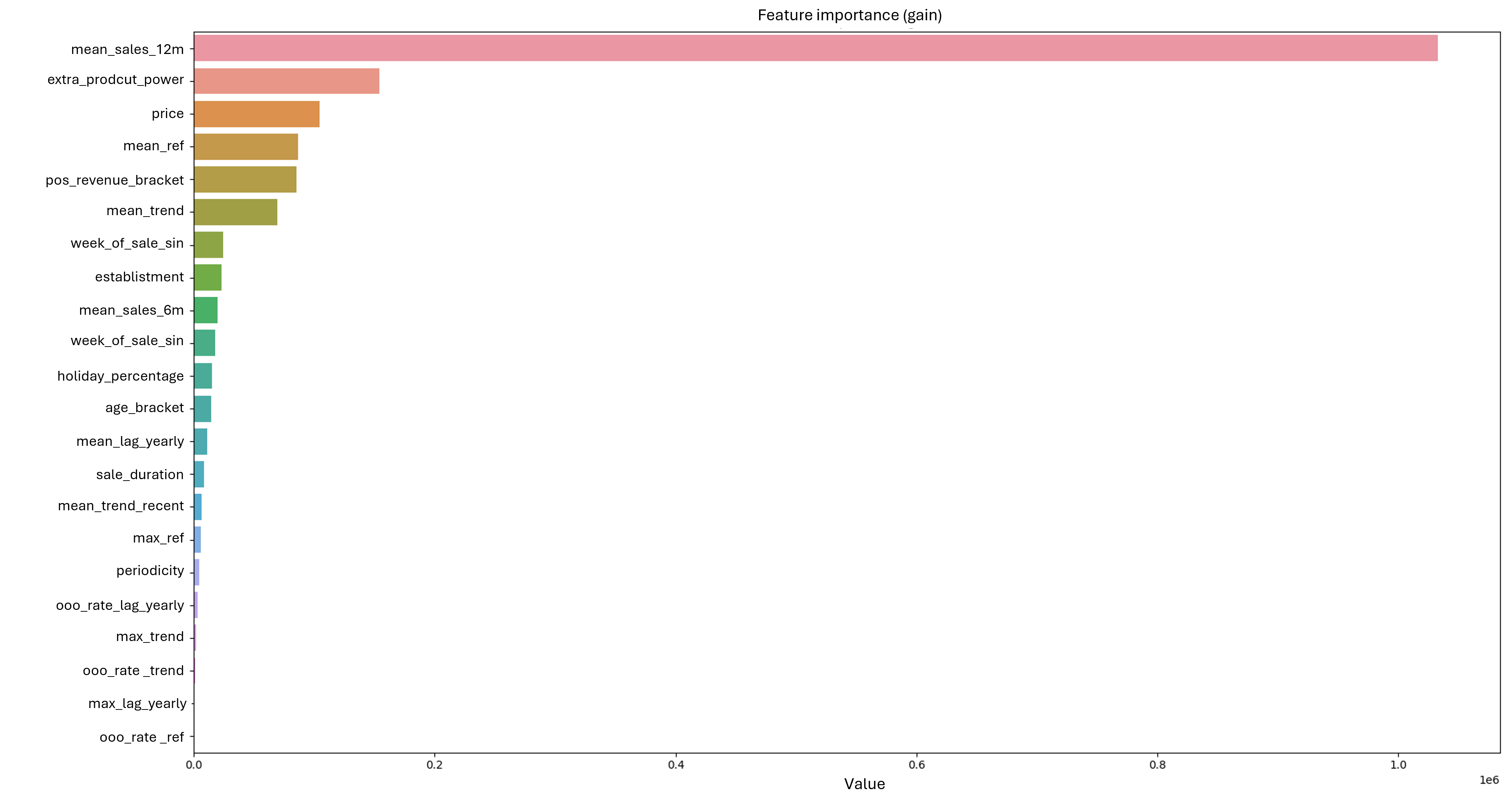}
  \centering
  \caption{Feature importance of a LightGBM\_log\_GCQR model.}.  \label{fig:feature_importance}
\end{figure}

\begin{figure}
  \centering
  \includegraphics[width=\linewidth]{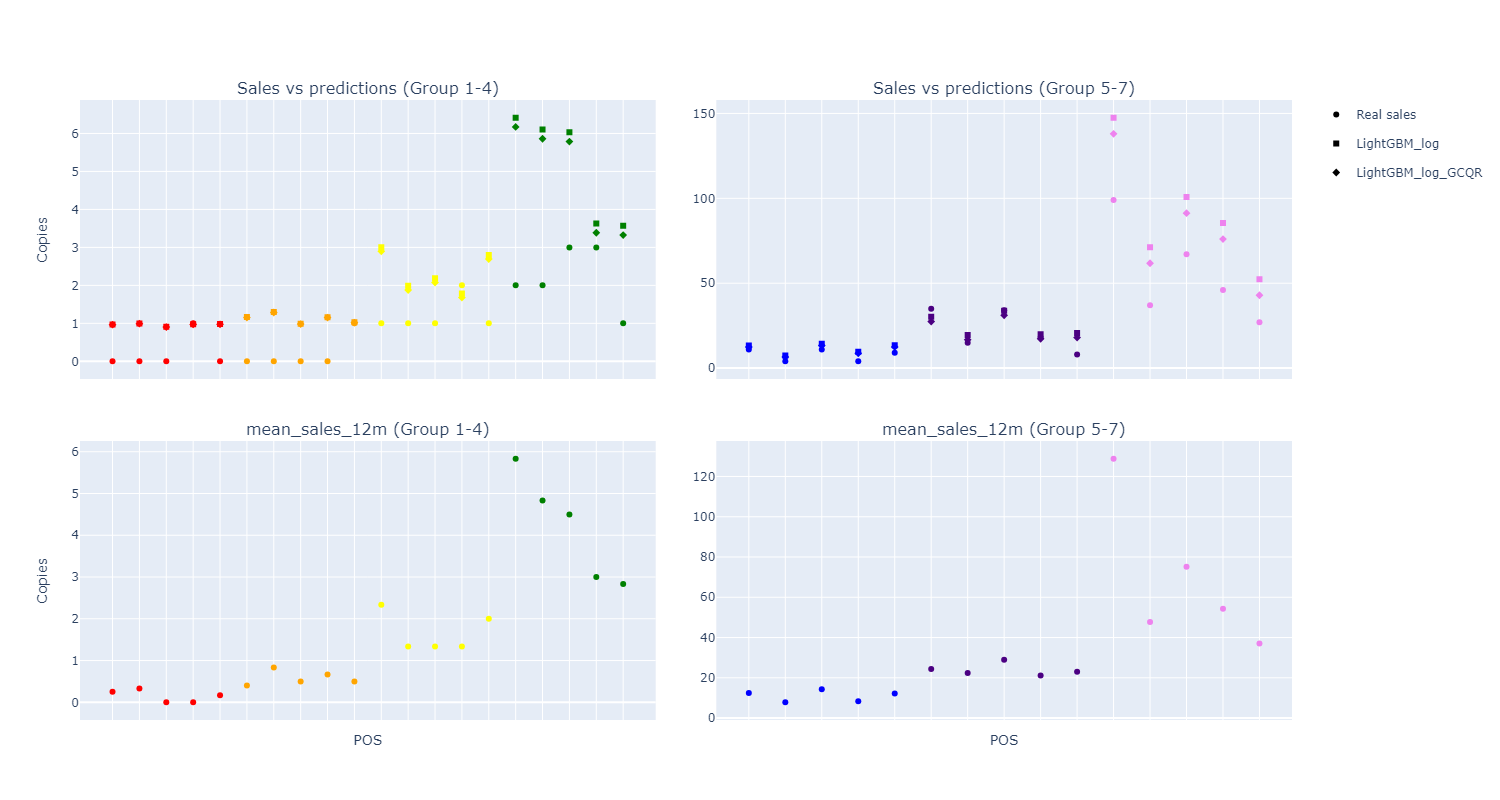}
  \centering
  \caption{Illustration of the differences between the predictions of LightGBM\_log and LightGBM\_log\_GCQR ($\alpha=0.85$). Top: the actual sales vs the predictions. Bottom: \texttt{mean\_sales\_12m}. Each color represents a group $G_g$. LightGBM\_log\_GCQR provided narrower intervals compared to LightGBM\_log, indicating improved prediction precision.}  \label{fig:correction_terms}
\end{figure}

\begin{figure}
  \centering
  \includegraphics[width=\linewidth]{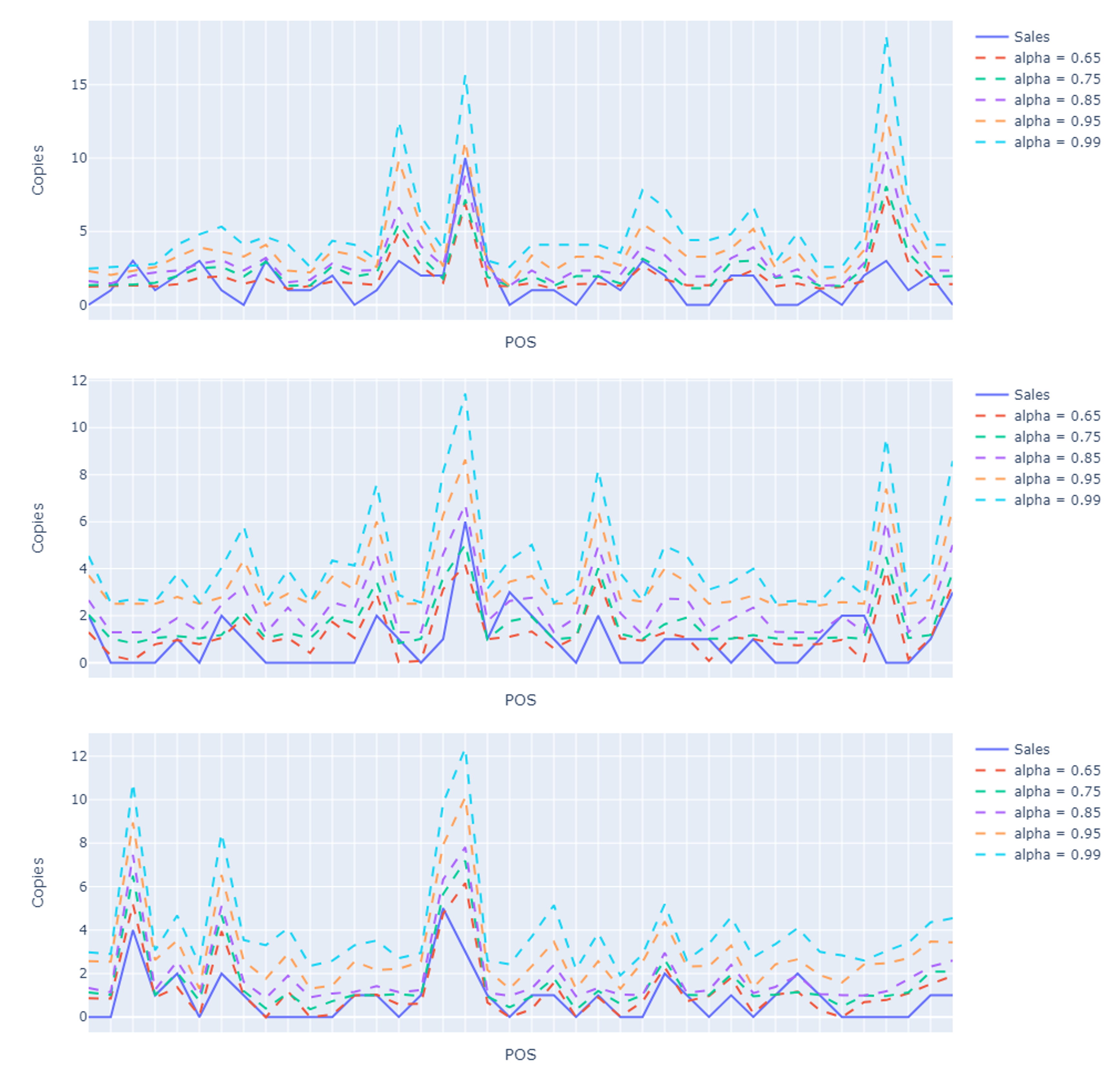}
  \centering
  \caption{Examples of the predictions of the demand sensing model for three issues of three different titles.}.  \label{fig:predictions}
\end{figure}

\rv{The results are shown in Table \ref{tab:demand_sensing_result}.  Note that we use the log scale only in the training phases of LightGBM\_log and LightGBM\_log\_GCQR. The calibration for LightGBM\_log\_GCQR and the evaluation of all the models are always conducted on the linear scale.  The poor performance of the Naive and S-Naive models can be attributed not only to their simplicity but also to their unsuitability for quantile regression. The superior performance of LightGBM\_linear, LightGBM\_log, and LightGBM\_log\_GCQR over LightGBM\_standard demonstrates the pertinence of the knowledge brought by the SMEs when providing the reference issues to the model. Figure \ref{fig:feature_importance} shows the feature importance of a LightGBM\_log\_GCQR model. The most important features of this model are \texttt{mean\_sale\_12m} (which reflects the average sales of the title at that POS), \texttt{extra\_product\_power}, \texttt{price}, \texttt{mean\_ref}, and \texttt{pos\_revenue\_bracket}. This aligns with business intuitions. Overall, LightGBM\_log\_GCQR models achieved the best quantile scores, supporting the choice of LightGBM\_log\_GCQR as the demand sensing model for the solution. The improvement of LightGBM\_log\_GCQR over LightGBM\_log is shown in Figure \ref{fig:correction_terms}. In general, LightGBM\_log\_GCQR provided narrower quantile intervals. Three examples of the predictions of this model are shown in Figure \ref{fig:predictions}. While using a high value of $\alpha$ helps avoid out-of-stock situations, it also creates oversupply.}

\begingroup
\begin{table*}[]
    \centering
    \begin{threeparttable}
    \begin{tabular}{l|*{7}c}
    Model & $\alpha$=0.65 & $\alpha$=0.75 & $\alpha$=0.85 & $\alpha$=0.95 &  $\alpha$=0.99 & Average \\
    \midrule 
    \midrule 
    Naive & 0.5426 & 0.5194 & 0.4961 & 0.4728 & 0.4635 & 0.4989 \\
    \midrule 
    S-Naive  & 0.5794 & 0.5488 & 0.5183 & 0.4877 & 0.4755 & 0.5219 \\
    \midrule 
    LightGBM\_standard  & 0.4171 & 0.3689 & 0.2927 & 0.1538 & 0.0442 & 0.2554 \\
    \midrule 
    LightGBM\_linear  & 0.4224 & 0.3833 & 0.3078 & 0.1571 & 0.0454 & 0.2632 \\
    \midrule 
    LightGBM\_log  & 0.4143 & 0.3691 & 0.2904 & 0.1549 & 0.0434 & 0.2544 \\
    \midrule 
    LigthGBM\_log\_GCQR (\textit{AthenIA}) & \textbf{0.4078} & \textbf{0.3602} & \textbf{0.2821} & \textbf{0.1499} & \textbf{0.0428} & \textbf{0.2486} \\
    \bottomrule
    \end{tabular}
    \end{threeparttable}
    \caption {Comparison of the performance of the demand sensing models.}
    \label{tab:demand_sensing_result}
\end{table*}
\endgroup

\subsection{Evaluation of the supply optimization method}
\label{subsec:eval_supply_optimization}

\rv{Table \ref{tab:optimization_result} shows the total supply, the profit, and the number of out-of-stock instances of different supply plans using the same setting as in Section \ref{subsec:eval_demand_sensing}. The ``quantile regression" plan is a supply plan that uses a quantile regression with $\alpha = 0.75$, then scales the supply proportionally to meet the total supply constraint.  Both the \textit{optimal distribution plan} and \textit{optimal supply plan} yielded better profits (101.91\% and 102.54\% of the actual profit, respectively) and fewer out-of-stock. Both these plans reduced the total supply and the number of out-of-stock instances at the same time, indicating that the estimations of the demand were more accurate. The \textit{optimal supply plan} tends to cut the supply to reduce the costs of oversupply. \rvv{Reducing the supply without sacrificing the profit is highly beneficial ecologically since it reduces the carbon footprints related to the production, distribution, and destruction of unsold copies. However, consistently using this supply plan might be counterproductive, as some level of oversupply can be beneficial. Maintaining a certain amount of unsold copies helps reduce the risk of out-of-stock and keeps the products visible to consumers. The total supply constraint $N_{total} \pm \Delta$ serves as an anchor to balance these considerations, ensuring that the supply level supports both profitability and product visibility.}}

\begingroup
\begin{table}[]
    \centering
    \begin{threeparttable}
    \begin{tabular}{l|*{3}c}
    Model & Total supply & Profit & Out-of-stock  \\
    \midrule 
    \midrule 
    Real & 100\% & 100\% & 100\%  \\
    \midrule 
    Quantile regression  & 99.01\% & 95.84\%  & 108.55\% \\
    \midrule 
    Optimal distribution  & 93.01\% &  101.97\% & 88.01\% \\
    \midrule 
    Optimal supply  & 90.08\%  & 102.54\%  & 95.42\% \\

    \bottomrule
    \end{tabular}
    \end{threeparttable}
    \caption {Comparison of the performance of the optimization methods.}
    \label{tab:optimization_result}
\end{table}
\endgroup

\section{Deployment}
\label{sec:deployment}

\textit{AthenIA} is a fully industrialized solution. The data are securely hosted in an Azure Data Lake, ensuring robust data management and security. Computational tasks, including data processing, model training and inference, and optimization, are executed on a Databricks cloud platform. To effectively manage and optimize the solution, we apply MLOps practices to ensure a robust and scalable lifecycle for \textit{AthenIA}. We refresh the solution weekly and deploy an automatic monitoring process to monitor the performance of the solution. To enhance the accessibility and usability of \textit{AthenIA}, we have built a web application to allow users to input product information (such as the price, the associated extra products, and the references), select and download the supply plans, and visualize the statistics of the issues (such as the sales, the sales rate, the total supply, the profit of historical issues). 

Every quarter, we hold a meeting with business stakeholders to review the performance of the business and assess the effectiveness of the current strategies and models. These sessions are crucial for defining actionable plans and making necessary adjustments to the products, the business strategies, and the solution itself. This process ensures that our strategies remain aligned with evolving business goals and market dynamics.

\section{Related work}
\label{sec:related_work}

\textbf{Magazine supply optimization}: There are limited published works on magazine supply optimization, with most studies focusing primarily on demand estimation, which is only a sub-task of the overall supply optimization process. In \cite{yu_support_2013}, the authors addressed the challenge of forecasting newspaper and magazine sales amidst the shift toward digital media consumption. They highlighted the inadequacy of traditional linear regression models, which often suffer from overfitting, and introduced Support Vector Regression (SVR) as an alternative. The authors estimated the sales based on the POS's geographical characteristics and the issue's selling period. A similar approach was studied in \cite{chu_predicting_2010}. In \cite{gencay_sales_2012} the authors utilized time series analysis to predict daily demand and minimized unsold copies. The research focused on customized forecasting models tailored to the unique characteristics of individual dealers, influenced by factors such as location and proximity to key institutions. Various forecasting techniques were compared, with a particular emphasis on a nonlinear approach for managing returns. The study highlighted the challenges of predicting sales on special days influenced by promotions and events, and concluded that ExponenTial Smoothing (ETS) yielded satisfactory results. However, that analysis was made on only one POS with sufficiently long historical data. Hence, it can not be generalized. In contrast, \textit{AthenIA} is an end-to-end industrialized supply optimization operating at a much larger scale.  

\textbf{Demand sensing}: \rv{Demand sensing is at the heart of supply chain management, especially in retail operations where accurate forecast of future demand can significantly enhance inventory management and profitability \cite{ren_demand_2020, fildes_retail_2022}. Various methods have been explored for demand sensing, each presents unique advantages and limitations. Statistical methods, such as Naive prediction, SARIMA \cite{shumway_arima_2017}, and exponential smoothing \cite{gardner_jr_exponential_1985}, have been widely used \cite{ren_fashion_2014, lin_predicting_2018} because of their simplicity and ease of implementation. These models are effective in capturing trends and seasonality in historical sales data and perform well for products with stable demand patterns. However, they often struggle with the high variability and unpredictability inherent in cases where demand can be influenced by numerous factors including trends, external events economic conditions. Recently, machine learning methods such as LightGBM \cite{ke_lightgbm_2017}, CatBoost \cite{prokhorenkova_catboost_2018}, DeepAR \cite{salinas_deepar_2019},  N-BEATS \cite{oreshkin_n-beats_2020}, and TFT \cite{lim_temporal_2020} have emerged as the state-of-the-art for demand sensing and demand forecasting \cite{ren_demand_2020, fildes_retail_2022}. These methods can handle non-linear relationships and complex interactions within the data, providing more accurate demand forecasts in dynamic environments. Each method, however, is suitable for specific use cases. In our case, the irregularity in the sales precludes the use of time series models such as DeepAR, N-BEATS, TFT. LightGBM is a more suitable choice.}

A critical aspect of effective demand sensing is feature engineering. The feature engineering step needs to adapt to the context and the objective of the use case. \textit{AthenIA} allows SMEs to incorporate their insights and prior knowledge of the products into the model. This human-in-the-loop approach enhances the model's ability to capture the nuances of the product characteristics and market dynamics, leading to more accurate and actionable forecasts.

\textbf{Conformalized quantile regression}: The idea of using different sets of \{training set, calibration set\} and training one prediction model for each set is similar to EnbPI \cite{xu_conformal_2023}. However, unlike EnbPI, which utilizes random sampling to generate different bootstrap sets, our approach involves deliberately choosing splits that are adaptive to each title.  The grouping step is to tackle the problem of heavy tail distribution of the demand. In contrast to \cite{romano_conformalized_2019}, our approach focuses on predicting a one-sided coverage interval.  Specifically, we aim to estimate the upper bound, represented by the $\alpha$th quantile of $D^j_{m,i}$. 

\textbf{Supply optimization}: \rv{The fixed inventory nature of our case precludes the use of traditional inventory management techniques such as Economic Order Quantity (EOQ), First-In-First-Out (FIFO), Last-In-First-Out (LIFO), Vendor-Managed Inventory (VMI), and Just-In-Time (JIT) \cite{aro-gordon_review_2016}. These methods rely on the ability to reorder stock, which is not applicable in scenarios where the inventory is predetermined and cannot be adjusted during the selling period. Modern algorithmic approaches for inventory management, such as dynamic programming \cite{bellman_applied_2015} and stochastic programming \cite{ross_introduction_2014}, offer advanced optimization capabilities but are computationally intractable for our case because of the extensive number of POSes. Reinforcement learning is a novel and promising technique in supply optimization \cite{bharti_reinforcement_2020}. However, it often requires a large amount of data and computational power, which can be limiting factors in practical applications. \textit{AthenIA} uses scenario simulation to leverage the precision of the demand sensing model. This approach enables the integration of various cost factors inherent in the complex structure of the magazine retail industry.}
\section{Conclusion}
\label{sec:conclusion}

In this paper, we presented \textit{AthenIA}, a supply optimization solution designed to address the intricate challenges faced by a French magazine publisher selling magazines at over 20,000 points of sale. We introduced a standardized yet agile pipeline for supply optimization, which stands out for its methodological rigor and flexibility. This pipeline is not only effective in the context of magazine distribution but also has the potential for application across other sectors thanks to its adaptability to diverse market conditions and business strategies.

\rv{A central innovation of \textit{AthenIA} is the development of a novel demand sensing approach that significantly enhances the integration of domain expert knowledge into the forecasting process. We designed a new group conformalized quantile regression technique to improve the performance of standard quantile regression. Furthermore, we introduced a tailored optimization model that maximizes the profit while considering different operational constraints, complex cost structures, and business strategies. }

\rv{\textit{AthenIA} is an example of the integration of advanced machine learning techniques with practical business insights to build a solution that is both innovative and applicable. The development of \textit{AthenIA} is also proof of the publisher's commitment to sustainability. The solution has been instrumental in helping significantly reduce the carbon footprint associated with over-supply in publishing. Through these concerted efforts, we are not only optimizing our supply chain but are also paving the way for a more sustainable and responsible operational model that other publishers can look to as a benchmark.}

\rv{In \textit{AthenIA}, Subject Matter Experts (SMEs) play a critical role in several steps of the supply optimization process. Their expertise is essential for accurately capturing the nuances of the product characteristics and market conditions. However, this reliance on human input introduces the potential for human bias and error. To mitigate these risks and use the solution effectively, it is crucial to maintain close collaboration with SMEs throughout the process. Future work may involve assessing the impact of the simplification that the demand is independent of the supply, which may not accurately reflect reality. Additionally, to further enhance the sustainability aspects of AthenIA, we could introduce carbon costs into the optimization equations. By incorporating these environmental costs, the solution would not only optimize for economic efficiency but also contribute to reducing the overall carbon footprint of the magazine supply chain. This holistic approach would align with broader sustainability goals and set a benchmark for responsible business practices in the publishing industry.} On the innovation side, a potential research direction is to leverage recent advances in natural language processing and multimodal processing to extract features from the contents of products and use them to estimate demand.

\section*{Acknowledgment}
\label{sec:acknowledgment}

This article represents the collective efforts of the \textit{AthenIA} team. We are grateful for their contributions in designing, building, and running the solution in production. Their dedication and expertise have been instrumental in the development of the solution. Special thanks to Vincent Rathelot for overseeing the operation of \textit{AthenIA}.

We also acknowledge the contributions of the Time Series team at Eki.Lab, particularly Eva Clergue's work during her internship at Ekimetrics.

\appendix
The details of the features used by the demand sensing models are described in Table \ref{tab:feature_engineering}.


\begingroup
\begin{table*}[htbp]
    \centering
    \begin{tblr}{
      colspec={|X|X|X[3]|}, row{1} = {c}, hlines,
    }
      Feature & Type & Description \\
        \texttt{establishment}   &  Categorical   &   Establishment type of the POS (e.g. supermarket, airport, shopping mall, camping hotel).   \\
        \texttt{pos\_revenue\_bracket} & Ordinal & Revenue bracket of the POS. \\
        \texttt{age\_bracket} &  Categorical   & Age bracket of the target customers. \\
        \texttt{periodicity} & Categorical & The periodicity of the title (e.g. weekly, monthly). \\
        \texttt{price} & Numerical & Price of the product \\
        \texttt{extra\_product\_power} & Ordinal &  Ranking of the extra-product based on the profit per unit of issues sold with that product.  \\
        \texttt{mean\_sales\_12m} & Numerical & Average sales per supplied issue over the last 12 months, lagged 3 months.  \\
        \texttt{mean\_sales\_6m} & Numerical  & Average sales per supplied issue over the last 6 months, lagged 3 months. \\
        \texttt{mean\_trend} & Numerical & Average sales per supplied issue over the three latest observed issues. \\
        \texttt{mean\_trend\_recent} & Numerical & Average sales per supplied issue over the two latest observed issues. \\
        \texttt{max\_trend} & Numerical & Maximum sales over the three latest observed issues. \\
        \texttt{oos\_rate\_trend} & Numerical & Number of out-of-stock divided by number of supplied issues, over the three latest observed issues. \\
        \texttt{mean\_ref} & Numerical & Average sales per supplied issue over the two references. \\
        \texttt{max\_ref} & Numerical & Maximum sales over the two references. \\
        \texttt{oos\_rate\_ref} & Numerical & Number of out-of-stock divided by number of supplied issues, over the two references. \\
        \texttt{mean\_lag\_yearly} & Numerical & Average sales per supplied issue over the 1-year lag and 2-year lag. \\
        \texttt{max\_lag\_yearly} & Numerical & Maximum sales over the 1-year lag and 2-year lag. \\
        \texttt{oos\_rate\_lag\_yearly} & Numerical & Number of out-of-stock divided by number of supplied issues, over the 1-year lag and 2-year lag. \\
        \texttt{selling\_duration} & Numerical & Selling duration of the issue (days). \\
        \texttt{holiday\_percentage} & Numerical & Number of public holidays during the selling period divided by the sale duration. \\
        \texttt{week\_of\_sale\_sin} & Numerical & The sinus of the cyclic encoding of the week number of the beginning of the selling period.  \\
        \texttt{week\_of\_sale\_cos} & Numerical & The cosinus of the cyclic encoding of the week number of the beginning of the selling period. \\
    
    \end{tblr}
    \caption{\raggedright Features used in the demand sensing model.}
    \label{tab:feature_engineering}
\end{table*}
\endgroup

\def\url#1{}
\bibliographystyle{IEEEtran}
\bibliography{references}

\begin{thebibliography}{10}
\providecommand{\url}[1]{#1}
\csname url@samestyle\endcsname
\providecommand{\newblock}{\relax}
\providecommand{\bibinfo}[2]{#2}
\providecommand{\BIBentrySTDinterwordspacing}{\spaceskip=0pt\relax}
\providecommand{\BIBentryALTinterwordstretchfactor}{4}
\providecommand{\BIBentryALTinterwordspacing}{\spaceskip=\fontdimen2\font plus
\BIBentryALTinterwordstretchfactor\fontdimen3\font minus \fontdimen4\font\relax}
\providecommand{\BIBforeignlanguage}[2]{{%
\expandafter\ifx\csname l@#1\endcsname\relax
\typeout{** WARNING: IEEEtran.bst: No hyphenation pattern has been}%
\typeout{** loaded for the language `#1'. Using the pattern for}%
\typeout{** the default language instead.}%
\else
\language=\csname l@#1\endcsname
\fi
#2}}
\providecommand{\BIBdecl}{\relax}
\BIBdecl

\bibitem{kaiser_magazine_2012}
U.~Kaiser and H.~C. Kongsted, ``Magazine {Companion} {Websites} and the {Demand} for {Newsstand} {Sales} and {Subscriptions},'' \emph{Journal of Media Economics}, vol.~25, no.~4, pp. 184--197, Oct. 2012, publisher: Routledge \_eprint: https://doi.org/10.1080/08997764.2012.729545.

\bibitem{noauthor_definition_nodate}
``Definition - {Domestic} duty on consumption of energy products / {TICPE} / {TIPP} / {TICPE} / {TIPP} {\textbar} {Insee}.''

\bibitem{noauthor_plastics_nodate}
``\BIBforeignlanguage{en}{Plastics own resource - {European} {Commission}}.''

\bibitem{gul_retail_2021}
E.~Gul, A.~Lim, and J.~Xu, ``\BIBforeignlanguage{en}{Retail {Store} {Layout} {Optimization} for {Maximum} {Product} {Visibility}},'' May 2021, arXiv:2105.09299 [math].

\bibitem{lu_developing_2015}
Y.~Lu and H.~Seo, ``Developing {Visibility} {Analysis} for a {Retail} {Store}: {A} {Pilot} {Study} in a {Bookstore},'' \emph{Environment and Planning B: Planning and Design}, vol.~0, pp. 0--0, Feb. 2015.

\bibitem{salinas_deepar_2019}
D.~Salinas, V.~Flunkert, and J.~Gasthaus, ``\BIBforeignlanguage{en}{{DeepAR}: {Probabilistic} {Forecasting} with {Autoregressive} {Recurrent} {Networks}},'' arXiv, Tech. Rep. arXiv:1704.04110, Feb. 2019, arXiv:1704.04110 [cs, stat] type: article.

\bibitem{oreshkin_n-beats_2020}
B.~N. Oreshkin, D.~Carpov, N.~Chapados, and Y.~Bengio, ``N-{BEATS}: {Neural} basis expansion analysis for interpretable time series forecasting,'' Feb. 2020, arXiv:1905.10437 [cs, stat].

\bibitem{lim_temporal_2020}
B.~Lim, S.~O. Arik, N.~Loeff, and T.~Pfister, ``\BIBforeignlanguage{en}{Temporal {Fusion} {Transformers} for {Interpretable} {Multi}-horizon {Time} {Series} {Forecasting}},'' arXiv, Tech. Rep. arXiv:1912.09363, Sep. 2020, arXiv:1912.09363 [cs, stat] type: article.

\bibitem{ke_lightgbm_2017}
G.~Ke, Q.~Meng, T.~Finley, T.~Wang, W.~Chen, W.~Ma, Q.~Ye, and T.-Y. Liu, ``{LightGBM}: {A} {Highly} {Efficient} {Gradient} {Boosting} {Decision} {Tree},'' in \emph{Advances in {Neural} {Information} {Processing} {Systems}}, vol.~30.\hskip 1em plus 0.5em minus 0.4em\relax Curran Associates, Inc., 2017.

\bibitem{prokhorenkova_catboost_2018}
L.~Prokhorenkova, G.~Gusev, A.~Vorobev, A.~V. Dorogush, and A.~Gulin, ``{CatBoost}: unbiased boosting with categorical features,'' \emph{Advances in neural information processing systems}, vol.~31, 2018.

\bibitem{laptev_time-series_2017}
N.~Laptev, J.~Yosinski, L.~E. Li, and S.~Smyl, ``Time-series extreme event forecasting with neural networks at uber,'' in \emph{International conference on machine learning}, vol.~34.\hskip 1em plus 0.5em minus 0.4em\relax sn, 2017, pp. 1--5.

\bibitem{kunz_deep_2023}
M.~Kunz, S.~Birr, M.~Raslan, L.~Ma, Z.~Li, A.~Gouttes, M.~Koren, T.~Naghibi, J.~Stephan, M.~Bulycheva, M.~Grzeschik, A.~Kekic, M.~Narodovitch, K.~Rasul, J.~Sieber, and T.~Januschowski, ``\BIBforeignlanguage{en}{Deep {Learning} based {Forecasting}: a case study from the online fashion industry},'' arXiv, Tech. Rep. arXiv:2305.14406, May 2023, arXiv:2305.14406 [cs] type: article.

\bibitem{koenker_regression_1978}
R.~Koenker and G.~Bassett~Jr, ``Regression quantiles,'' \emph{Econometrica: journal of the Econometric Society}, pp. 33--50, 1978, iSBN: 0012-9682 Publisher: JSTOR.

\bibitem{romano_conformalized_2019}
Y.~Romano, E.~Patterson, and E.~J. Candes, ``\BIBforeignlanguage{en}{Conformalized {Quantile} {Regression}},'' arXiv, Tech. Rep. arXiv:1905.03222, May 2019, arXiv:1905.03222 [stat] type: article.

\bibitem{shafer_tutorial_2008}
G.~Shafer and V.~Vovk, ``\BIBforeignlanguage{en}{A {Tutorial} on {Conformal} {Prediction}},'' \emph{\BIBforeignlanguage{en}{Journal of Machine Learning Research}}, vol.~9, no.~3, 2008.

\bibitem{angelopoulos_gentle_2022}
A.~N. Angelopoulos and S.~Bates, ``\BIBforeignlanguage{en}{A {Gentle} {Introduction} to {Conformal} {Prediction} and {Distribution}-{Free} {Uncertainty} {Quantification}},'' arXiv, Tech. Rep. arXiv:2107.07511, Dec. 2022, arXiv:2107.07511 [cs, math, stat] type: article.

\bibitem{yu_support_2013}
X.~Yu, Z.~Qi, and Y.~Zhao, ``\BIBforeignlanguage{en}{Support {Vector} {Regression} for {Newspaper}/{Magazine} {Sales} {Forecasting}},'' \emph{\BIBforeignlanguage{en}{Procedia Computer Science}}, vol.~17, pp. 1055--1062, Jan. 2013.

\bibitem{chu_predicting_2010}
M.~L.~Z. Chu, F.~Fan, and Y.~Peng, ``Predicting {Magazine} {Sales} {Using} {Machine} {Learning},'' \emph{url: http://citeseerx. ist. psu. edu/viewdoc/download}, p.~17, 2010.

\bibitem{gencay_sales_2012}
I.~Gencay, A.~Baris, and T.~Mujgan, ``Sales forecasting system for newspaper distribution companies in {Turkey},'' \emph{Pakistan Journal of Statistics and Operation Research}, pp. 685--699, 2012, iSBN: 2220-5810 Publisher: College of Statistical and Actuarial Sciences.

\bibitem{ren_demand_2020}
S.~Ren, H.-L. Chan, and T.~Siqin, ``\BIBforeignlanguage{en}{Demand forecasting in retail operations for fashionable products: methods, practices, and real case study},'' \emph{\BIBforeignlanguage{en}{Annals of Operations Research}}, vol. 291, no. 1-2, pp. 761--777, Aug. 2020.

\bibitem{fildes_retail_2022}
R.~Fildes, S.~Ma, and S.~Kolassa, ``\BIBforeignlanguage{en}{Retail forecasting: {Research} and practice},'' \emph{\BIBforeignlanguage{en}{International Journal of Forecasting}}, vol.~38, no.~4, pp. 1283--1318, Oct. 2022.

\bibitem{shumway_arima_2017}
R.~H. Shumway and D.~S. Stoffer, ``\BIBforeignlanguage{en}{{ARIMA} {Models}},'' in \emph{\BIBforeignlanguage{en}{Time {Series} {Analysis} and {Its} {Applications}: {With} {R} {Examples}}}, R.~H. Shumway and D.~S. Stoffer, Eds.\hskip 1em plus 0.5em minus 0.4em\relax Cham: Springer International Publishing, 2017, pp. 75--163.

\bibitem{gardner_jr_exponential_1985}
E.~S. Gardner~Jr, ``Exponential smoothing: {The} state of the art,'' \emph{Journal of forecasting}, vol.~4, no.~1, pp. 1--28, 1985, iSBN: 0277-6693 Publisher: Wiley Online Library.

\bibitem{ren_fashion_2014}
S.~Ren, T.-M. Choi, and N.~Liu, ``Fashion sales forecasting with a panel data-based particle-filter model,'' \emph{IEEE Transactions on Systems, Man, and Cybernetics: Systems}, vol.~45, no.~3, pp. 411--421, 2014, iSBN: 2168-2216 Publisher: IEEE.

\bibitem{lin_predicting_2018}
L.~Lin, Z.~He, and S.~Peeta, ``Predicting station-level hourly demand in a large-scale bike-sharing network: {A} graph convolutional neural network approach,'' \emph{Transportation Research Part C: Emerging Technologies}, vol.~97, pp. 258--276, 2018, iSBN: 0968-090X Publisher: Elsevier.

\bibitem{xu_conformal_2023}
C.~Xu and Y.~Xie, ``\BIBforeignlanguage{en}{Conformal prediction for time series},'' arXiv, Tech. Rep. arXiv:2010.09107, Feb. 2023, arXiv:2010.09107 [stat] type: article.

\bibitem{aro-gordon_review_2016}
S.~Aro-Gordon and J.~Gupte, ``Review of modern inventory management techniques,'' \emph{The Global Journal of Business and Management}, vol.~1, pp. 1--22, Sep. 2016.

\bibitem{bellman_applied_2015}
R.~E. Bellman and S.~E. Dreyfus, \emph{\BIBforeignlanguage{en}{Applied {Dynamic} {Programming}}}.\hskip 1em plus 0.5em minus 0.4em\relax Princeton University Press, Dec. 2015, google-Books-ID: ZgbWCgAAQBAJ.

\bibitem{ross_introduction_2014}
S.~M. Ross, \emph{Introduction to stochastic dynamic programming}.\hskip 1em plus 0.5em minus 0.4em\relax Academic press, 2014.

\bibitem{bharti_reinforcement_2020}
S.~Bharti, D.~S. Kurian, and V.~M. Pillai, ``Reinforcement learning for inventory management,'' in \emph{Innovative {Product} {Design} and {Intelligent} {Manufacturing} {Systems}: {Select} {Proceedings} of {ICIPDIMS} 2019}.\hskip 1em plus 0.5em minus 0.4em\relax Springer, 2020, pp. 877--885.

\end{thebibliography}

\end{document}